\RequirePackage{fix-cm}
\documentclass[twocolumn]{mysvjour}          
\smartqed  
\usepackage{graphicx}
\begin{document}

\title{Development of a Real-time Colorectal Tumor Classification System for Narrow-band Imaging zoom-videoendoscopy}
\subtitle{}


\author{Tsubasa Hirakawa \and Toru Tamaki \and  Bisser Raytchev \and Kazufumi Kaneda \and Tetsushi Koide \and Shigeto Yoshida \and Hiroshi Mieno \and Shinji Tanaka
}


\institute{T. Hirakawa \at
           Corresponding author \\
           \email{hirakawa@eml.hiroshima-u.ac.jp}
           \and
           T. Hirakawa \and T. Tamaki \and B. Raytchev \and K. Kaneda \at 
              Department of Information Engineering, Graduate School of Engineering, Hiroshima University, 1-4-1 Kagamiyama, Higashi-Hiroshima, Hiroshima 739-8527, Japan
           \and
           T. Koide \at
              Research Institute for Nanodevice and Bio Systems (RNBS), Hiroshima University, 1-4-2 Kagamiyama, Higashi-Hiroshima 739-8527, Japan
           \and
           S. Yoshida \and H. Mieno \at
              Department of Gastroenterology, Hiroshima General Hospital of West Japan Railway Company, 3-1-36 Futabanosato, Higashiku, Hiroshima 732-0057, Japan
           \and
           S. Tanaka \at
              Department of Endoscopy, Hiroshima University Hospital,1-2-3 Kasumi, Minami-ku, Hiroshima 734-8551, Japan
}

\date{}

\maketitle

\begin{abstract}
Colorectal endoscopy is important for the early detection and treatment of colorectal cancer and is used worldwide. A computer-aided diagnosis (CAD) system that provides an objective measure to endoscopists during colorectal endoscopic examinations would be of great value. In this study, we describe a newly developed CAD system that provides real-time objective measures. Our system captures the video stream from an endoscopic system and transfers it to a desktop computer. The captured video stream is then classified by a pretrained classifier and the results are displayed on a monitor. The experimental results show that our developed system works efficiently in actual endoscopic examinations and is medically significant.
\end{abstract}

\section*{Introduction}
\label{intro}
\begin{figure*}[tb]
\centering
\includegraphics[width=0.75\linewidth]{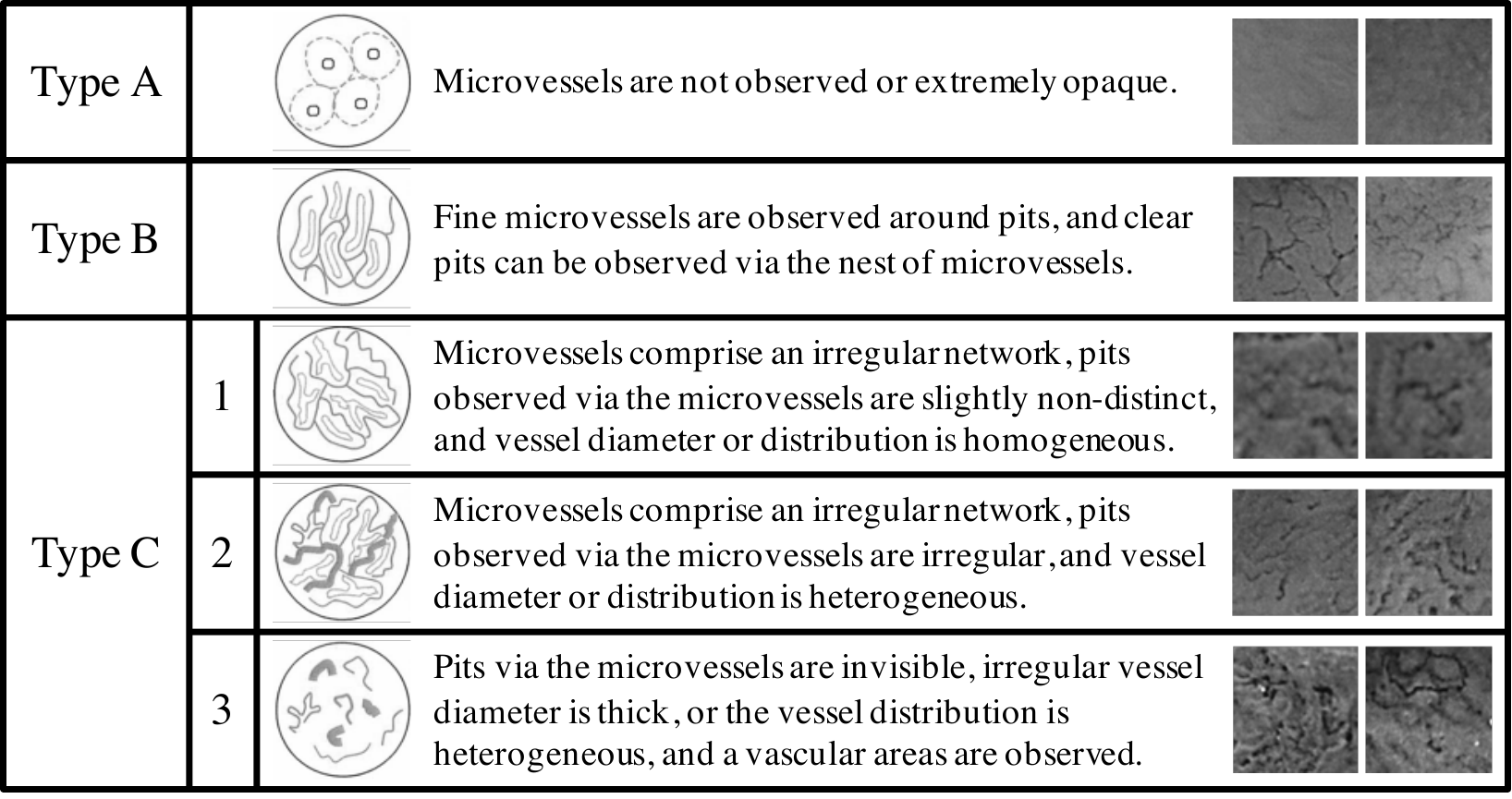}
\caption{NBI magnification findings \cite{Kanao2009}.}
\label{fig:NBImag}
\end{figure*}

Endoscopic examinations for colorectal cancer, called colorectal endoscopies or colonoscopies, are used for the detection and diagnosis of early-stage colorectal cancer. During an endoscopic examination of the colon, endoscopists observe a video stream of the tumor, which is captured by the charge-coupled device (CCD) of an endoscope and is displayed on a monitor, to determine whether treatment and resection of the tumor are necessary \cite{Takemura2012}. Owing to the recent development of zoom-videoendoscopes with large magnification factors (up to 100), the endoscopists can visually inspect tumors in great detail. A chromoendoscopy may take a long time because of the cost of spraying and vacuuming different dyes to enhance the microvessel patterns of the mucosal surface. Narrow-band imaging (NBI) systems \cite{Gono2004,Machida2004,Sano2006} have been used to shorten the duration of the endoscopic examination. Using two light sources whose different wavelengths are absorbed by the hemoglobin in the blood vessels, NBI provides enhanced images of microvessel structures. To use images of the microvessel patterns of colon tumors taken by NBI zoom-endoscopes for visual inspection, an NBI magnification findings \cite{Oba2010,Kanao2009} was proposed by the Hiroshima University Hospital, which categorizes tumors into types A, B, and C, with type C further sub-classified into types C1, C2, and C3 on the basis of the enhanced micro-vessel structures (see Figure \ref{fig:NBImag}).

However, the intraobserver/interobserver variability \cite{Oba2010,Meining2004,Mayinger2006} shows that a diagnosis made by visual inspection on the basis of the NBI magnification findings could be subjective and the accuracy of the diagnosis depends on the experience of the endoscopist. Therefore, developing a computer-aided diagnosis (CAD) system to support visual inspections would be beneficial and would provide an objective measure of the status of colon tumors in each video frames of the videoendoscope. Such a system could make the diagnosis more objective and skill-independent. From the viewpoint of supporting endoscopists, providing such a real-time objective measure during the examination as opposed to after the examination would be beneficial \cite{Takemura2012} because endoscopists can use it to diagnose the condition of the tumor during examinations.

One possible and promising method to achieve this goal is to get a part of each video frame classified by a pretrained classifier and to show the classification results on a monitor during each video frame. Our previous prototype CAD system \cite{Tamaki2013MedIA} used the bag-of-visual words (BoVW) framework \cite{Csurka2004,Nowak2006,Lazebnik2006} with a densely sampled scale-invariant feature transform (SIFT) \cite{Lowe1999,Lowe2004,FeiFei2005,Jurie2005,Herve2009}, hierarchical k-means clustering \cite{Nister2006}, and a support vector machine (SVM) classifier \cite{Stenwart2008,Scholkopf2002,Cristianini2000,Vapnik1998}. This system is offline and classifies image patches trimmed from NBI endoscopic video frames into two or three categories. On the basis of the NBI magnification findings, image patches are classified into type A (corresponding to non-neoplastic lesions) or other types (neoplastic lesions) for a two-category classification or into type A, B, or C3 for a three-category classification. Several studies \cite{Hirakawa2013EMBC,Hirakawa2013ICIP,Hirakawa2016} have extended this CAD system to accommodate video frames from the NBI zoom-videoendoscope, with the system classifying a part of each endoscopic video frame. As an objective measure of the status of the tumor, classification probabilities are computed for each frame. However, these extensions have only been applied to offline videos (i.e., reading a stored movie file and processing each video frame) and have not yet been used or validated in actual clinical examinations.

In this study, we describe a newly developed CAD system that provides a real-time objective measure to endoscopists during examinations. Our system captures the online video stream from a videoendoscope via a video capture board-equipped desktop computer, converts the video format, and classifies each frame using a pretrained patch-based classifier \cite{Tamaki2013MedIA}. Finally, the obtained classification results are displayed on a monitor. In the next section, we describe the CAD system design that enables the system to be mobile and compatible with different endoscopic systems in a hospital, the software part of the system, and the patch-based-classifier. We have specified the following three requirements that the developed CAD system must satisfy: mobility, high frame rate, and medical significance. Experimental results show that our system has a mobility and a sufficiently fast processing speed. In terms of medical significance, a requirement for the accuracy of the real-time assessment by endoscopists has been provided by a medical society \cite{Rex2011}, which is discussed in the Results and Discussion section. Following six months of clinical case studies of the developed CAD system in actual endoscopic examinations at the Hiroshima University Hospital, we showed that it is our system allows nonexpert endoscopists to diagnose with sufficient accuracy needed to meet the specified requirements.

\begin{figure*}[tb]
\centering
\includegraphics[width=0.9\linewidth]{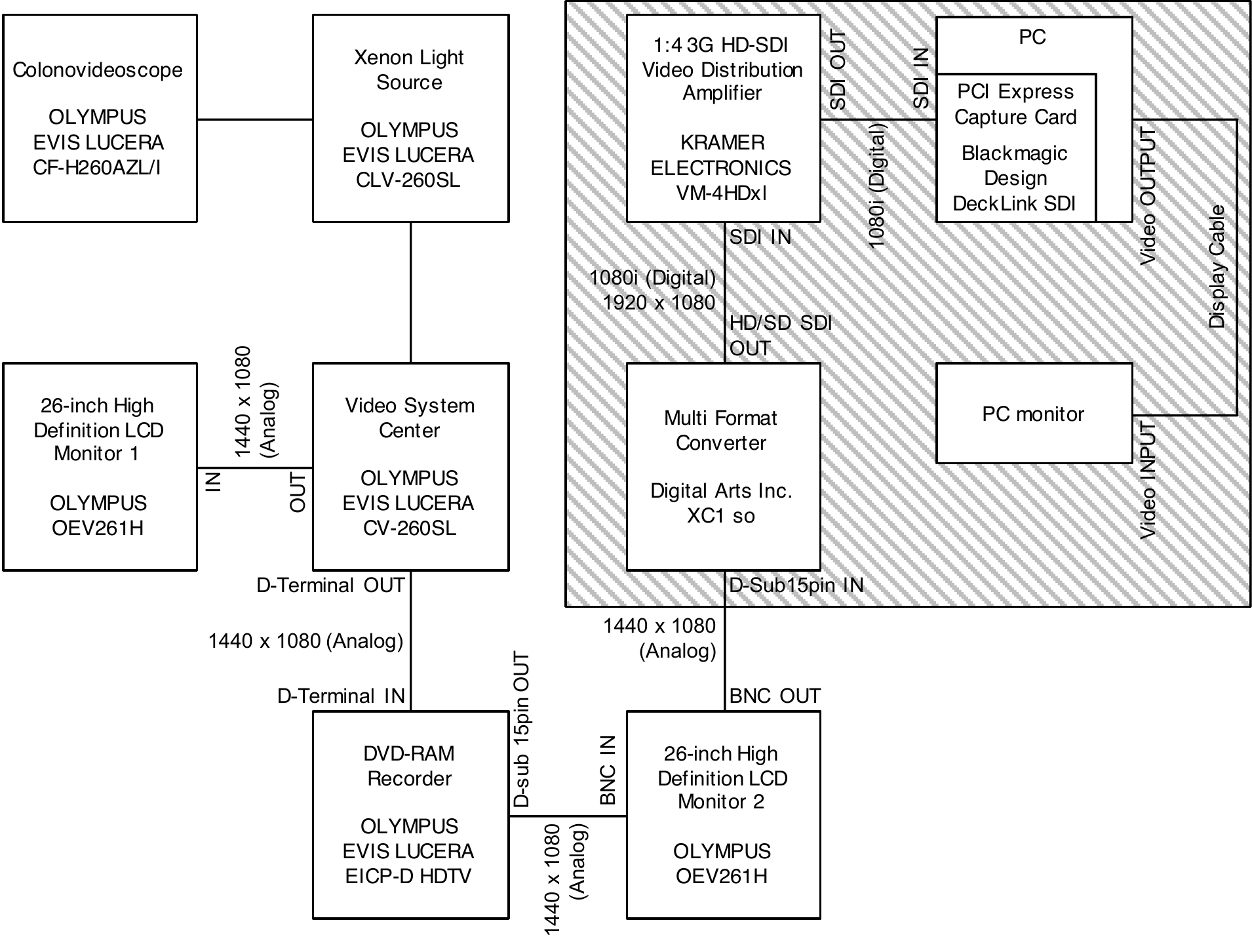}
\caption{An illustration of the system configuration of the endoscopic system (Olympus EVIS LUCERA) and our developed CAD system (gray shaded area).}
\label{fig:system_old}
\end{figure*}
\section*{Materials and Methods}

\begin{figure*}[tb]
\centering
\includegraphics[width=0.9\linewidth]{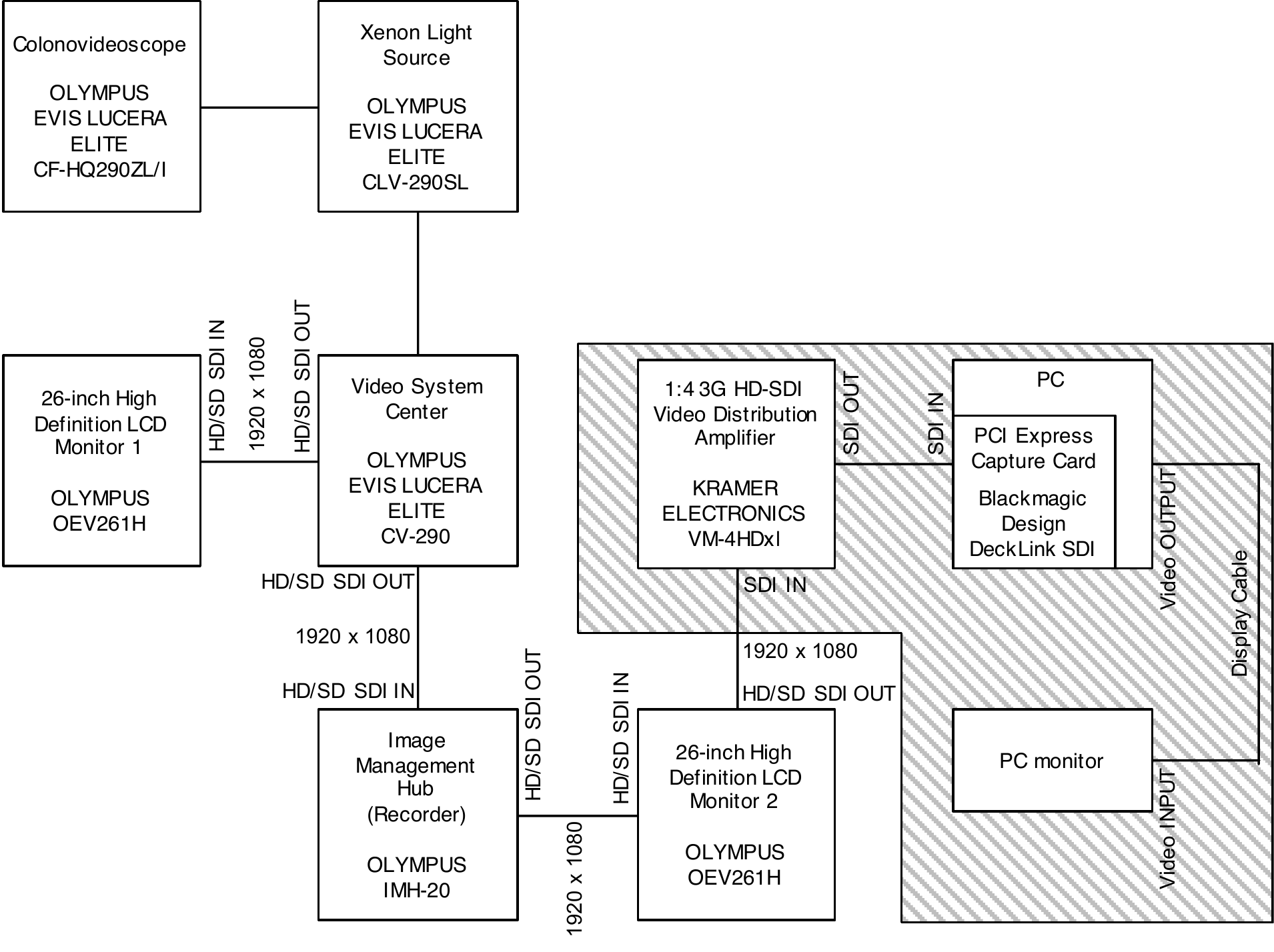}
\caption{An illustration of the system configuration of the endoscopic system (Olympus EVIS LUCERA ELITE) and our developed CAD system (gray shaded area).}
\label{fig:system_new}
\end{figure*}

\begin{figure*}[tb]
\centering
\includegraphics[width=0.8\linewidth]{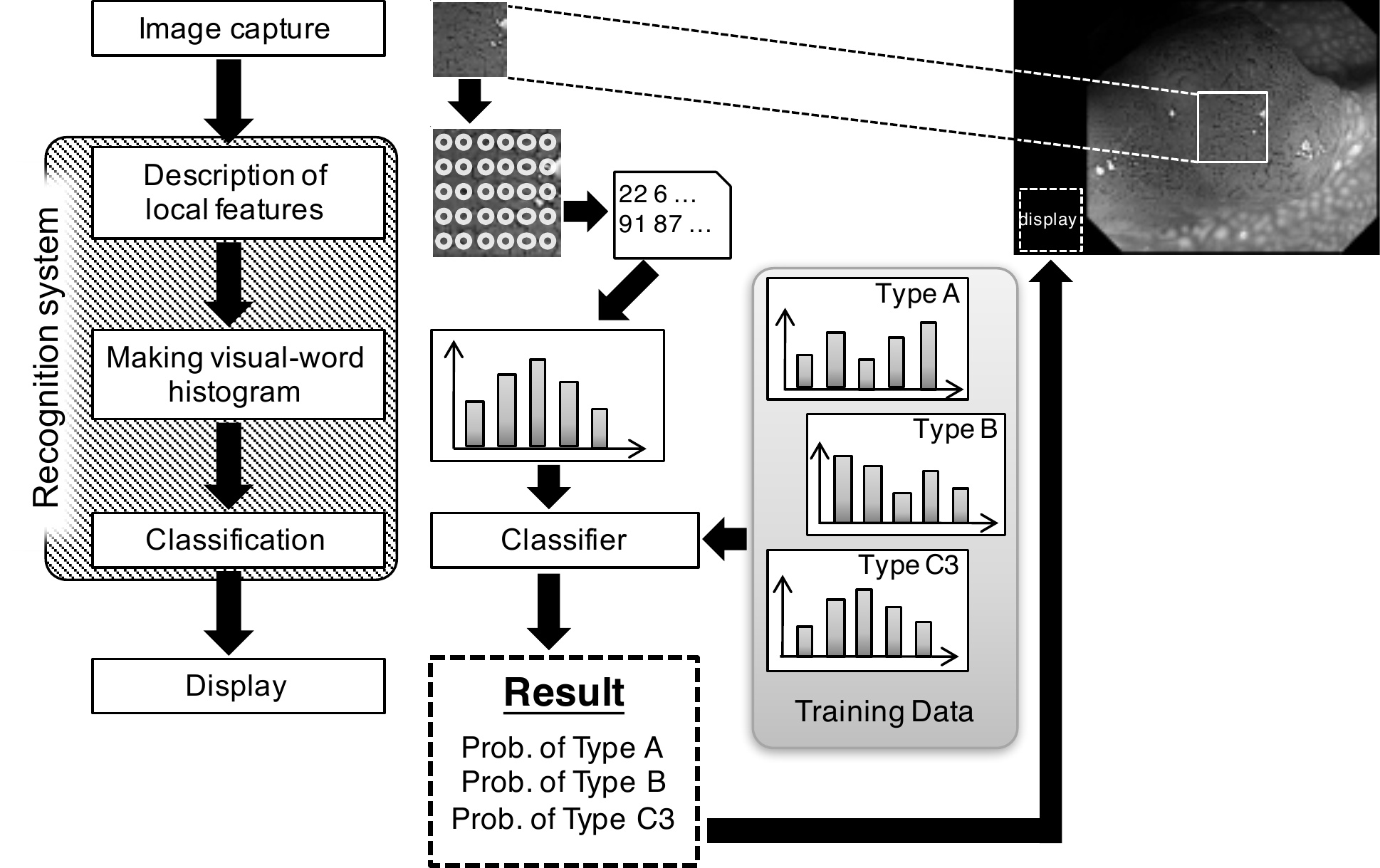}
\caption{Overview of video frame classification \cite{Tamaki2013MedIA}.}
\label{fig:recogsystem}	
\end{figure*}

\subsection*{Design of the video stream capturing system}
Herein, we describe the system design of the developed CAD system. There are two requirements the system needs to meet. First, we cannot modify the current configurations of the endoscopic systems that are regularly used in the hospital for our developed system and experiments. In this study, we used two different endoscopic systems: Olympus EVIS LUCERA (Figure \ref{fig:system_old}) and Olympus EVIS LUCERA ELITE (Figure \ref{fig:system_new}). These endoscopic systems have been configured and adjusted for regular examinations. Changing the configuration, e.g., switching the cables to intercept video streams, could cause the actual clinical flow between the endoscopy and the storage for medical data to stop. Second, the developed system must be able to deal with the two different endoscopic systems. In general, there are different endoscopic devices of different generations and types in different hospitals or even in a single hospital; hence, the developed system should have good mobility within the hospital. Keeping this in mind, we designed our system to capture the video stream from the videoendoscopes in such a way that simply attaching or detaching the connector for the video branch would be sufficient for operation.

The structure of the developed CAD system for one of the endoscope systems (Olympus Optical Co, Ltd; EVIS LUCERA) is shown in Figure \ref{fig:system_old}. The video stream captured by a colonovideoscope (Olympus Optical Co, Ltd; CF-H260AZL/I) with a xenon light source (Olympus Optical Co, Ltd; CLV-260SL) is sent to the video system center (Olympus Optical Co, Ltd; CV-260SL) to be processed. The processed stream is transferred to a digital versatile disc (DVD)- random access memory (RAM) recorder (Olympus Optical Co, Ltd; EICP-D HDTV) and is then passed to the second monitor (Olympus Optical Co, Ltd; OEV261H). The analog (RGB) video stream from the DVD-RAM recorder is displayed on the second monitor.

To capture the video stream for our system, we used the bypassed video stream from the second monitor. Because the video stream is analog in all the connections of this endoscopic system, we needed to convert it to a digital video stream using a multiformat video converter (XC1 co; Digital Arts Inc.); this conversion degrades the image quality. We then split the converted digital video stream by inserting a distribution amplifier (VM-4HDxl; Kramer Electronics Ltd.) and transferred it to a desktop computer (Intel Core i7-4770 3.4 GHz CPU with 16 GB memory, Microsoft Windows 7 Home Premium SP1 64bit) equipped with a peripheral component interconnect (PCI) express video capture card (DeckLink SDI; Blackmagic Design Pty. Ltd.).

Figure \ref{fig:system_new} shows the structure of the developed CAD system for another endoscope system (Olympus Optical Co, Ltd; EVIS LUCERA ELITE) comprising a colonovideoscope (Olympus Optical Co, Ltd; CF-HQ290 ZL/I), a xenon light source (Olympus Optical Co, Ltd; CLV-290SL), a video system center (Olympus Optical Co, Ltd; CV-290), a recorder (Olympus Optical Co, Ltd; IMH-20), and monitors (Olympus Optical Co, Ltd; OEV261H).

Similar to the first system, which is shown in Figure \ref{fig:system_old}, the video stream bypassed at the monitor is branched by the video distribution amplifier and is then transferred into the PCI express video capture card on the desktop computer. Note that a video converter is not required because the video stream is digital in all the connections of this endoscopic system.

At the end of the flow in the developed CAD system, the video stream is captured using the software development kit (SDK) of the capture card (DeckLink SDK 10.0; Blackmagic Design Pty. Ltd.). Then, the video frame is converted from YUV422 format to RGB format and is stored using an image-processing library (OpenCV 3.0 developer version; OpenCV.org) \cite{Bradski2000}. This color conversion is necessary because the digital video stream is usually represented in YUV color space whereas the software usually uses RGB color space for image processing. After the color conversion, the converted RGB video frame is passed to the patch-based classifier to compute the results. This is described in the following subsection.

\subsection*{Endoscopic video frame classification system}
An overview of the online classification of the video frames is shown in Figure \ref{fig:recogsystem}. The region-of-interest (ROI) is set to a rectangular patch at the center of the frame of the videoendoscope (a white rectangle inside the video frame in the upper right of Figure \ref{fig:recogsystem}). Then, densely sampled SIFT descriptors \cite{Lowe1999,Lowe2004,FeiFei2005,Jurie2005,Herve2009} computed on a regular grid of 5 pixels with two different scales (5 and 7 pixels) are extracted in the ROI. The extracted SIFT descriptors are represented as a histogram of the visual words (representative SIFT features) computed by hierarchical k-means clustering \cite{Nister2006}. Each bin of this histogram is linearly scaled with a fixed factor (determined at the training phase; see below) because the scaling is well-known to affect the classification performance. This histogram feature (usually called the BoVW histogram feature) is then classified by a pretrained linear SVM classifier \cite{Stenwart2008,Scholkopf2002,Cristianini2000,Vapnik1998} to obtain the classification probabilities for each category. These results (probabilities) are finally displayed on a monitor by superimposing the results onto the captured video frame. In this implementation, we used VLFeat 0.9.18 \cite{Vedaldi2010} for extracting the SIFT descriptors and for hierarchical k-means clustering, and LIBSVM 2.91 \cite{Chang2011} for the linear SVM classifier. This online classification was developed under an integrated development environment (IDE; Visual Studio 2012; Microsoft Corp.) and written in C++.

\begin{figure}[tb]
\centering
\includegraphics[width=\linewidth]{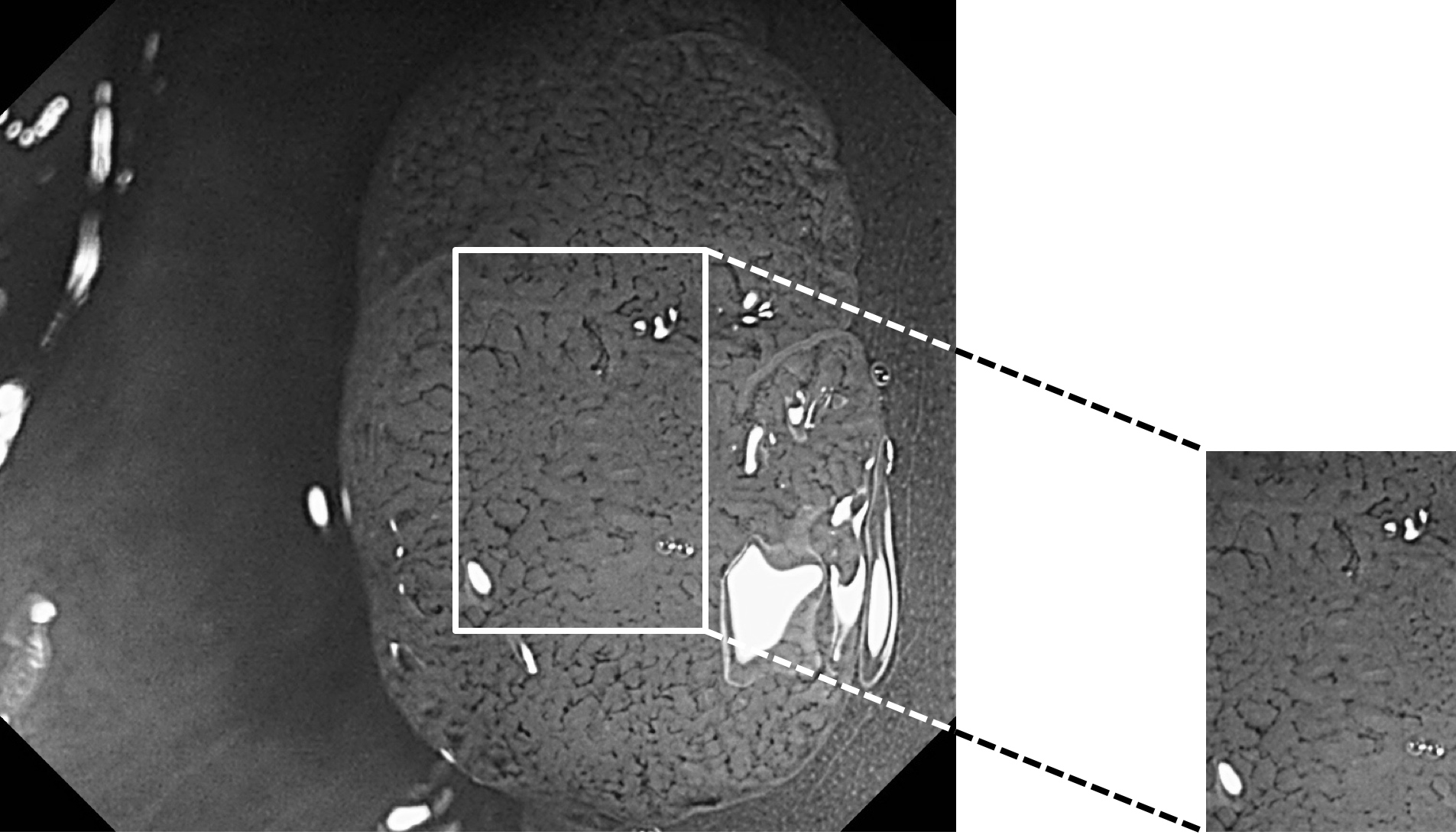}
\caption{Training sample construction by trimming a rectangle (right) from an NBI videoendoscope image (left).}
\label{fig:trim}
\end{figure}

To train the SVM classifier, we collected training samples as follows. A still image of a video frame was trimmed (see Figure \ref{fig:trim}) by endoscopists into a rectangular patch that contained typical microvessel structures. Then, labels were assigned to the image patches by endoscopists. Currently, we have a dataset of 2247 NBI image patches (Type A: 504, Type B: 847, Type C1: 257, Type C2: 57, and Type C3: 582) for two- or three-category classifications. Note that all these trimmed images were collected and all experiments of this current study were performed at the Hiroshima University Hospital. The guidelines of the Hiroshima University ethics committee were followed, and informed consent was obtained from the patients and their families.

The classifier-training phase was conducted offline before the online classification. We trained a linear SVM classifier using the training samples mentioned above, and all the necessary parameters of the SVM classifier were stored in a file. During the online classification phase, this file was loaded and used for the SVM classification (therefore, we call it a pretrained classifier). We used the same procedure to compute the BoVW histogram feature in both the training and online classification phases. First, densely sampled SIFT descriptors were extracted from the training samples and the BoVW histogram features were constructed. Each bin of the histogram was scaled linearly so that the range of values of the bin was [$-1$,$1$] throughout for the training samples; we then stored the scaling factor. Subsequently, a linear SVM was trained with the BoVW histogram features of the training samples for different values of the penalty parameter using a 5-fold cross validation. The values that gave the best performance were selected and stored. Readers interested in the details of the classification and training scheme can refer to Ref. \cite{Tamaki2013MedIA}. This training phase was implemented on a different desktop computer (Intel Xeon CPU E5-2620 with 128 GB memory, Ubuntu 14.04LTS; Canonical Ltd.) with several different codes written in C++ using the same libraries as the ones used in online classification. These codes were integrated with Bash and Perl 5.18 scripts for parallel processing.

 possible problem that might arise if our system is continuously used for several years is the discrepancy between the training samples and test samples due to a difference in the endoscopic devices. University hospitals are likely to replace endoscopic devices after a specific period of time. In that case, we need to collect as many training samples as possible for the new device to obtain acceptable classification results by training classifiers with the newly collected samples. However, it is impractical to collect a large number of training samples in a short period of time for each endoscope. To overcome this problem, we use a transfer learning-based learning approach proposed by Sonoyama et al. \cite{Sonoyama2015,Sonoyama2016} that trains a classifier with samples of the new endoscope by reusing (or transferring) the training samples from the old one. This enables us to maintain the classification performance without recollecting training samples when switching our system to a new endoscope.

\begin{figure}[tb]
\centering
\includegraphics[width=\linewidth]{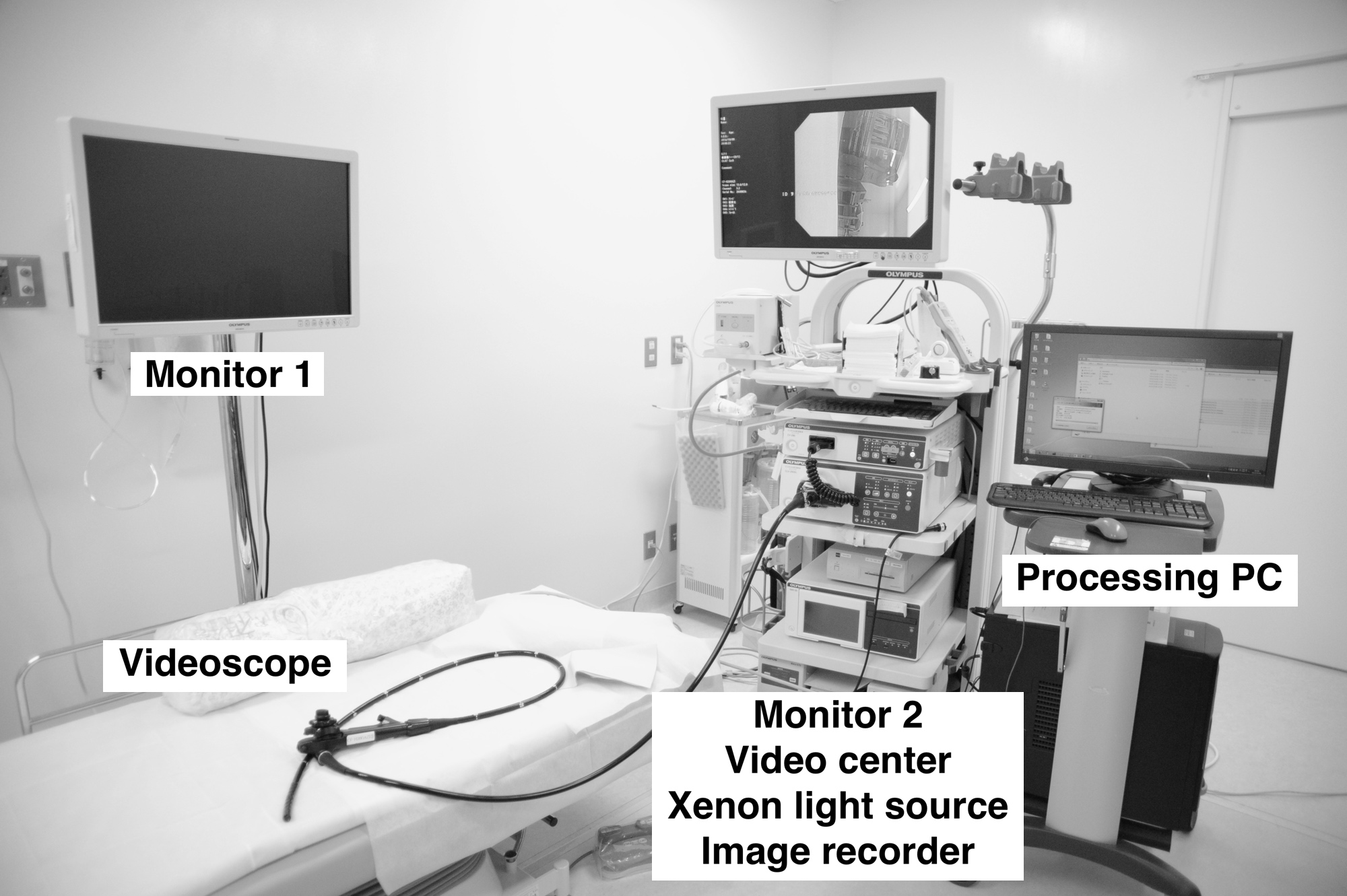}
\caption{The endoscopic system and the developed system.}
\label{fig:overview_system}
\end{figure}

\section*{Results and Discussion}

A snapshot of the endoscopic system and the developed CAD system is shown in Figure \ref{fig:overview_system}. The developed system consists of a desktop computer, a monitor, a keyboard, and a mouse arranged in a single mobile rack. The use of the PCI video capture card makes the system look slightly large; however, it is not difficult to move the rack from one endoscope to another. To make it smaller, we could develop a system on a laptop computer using a small video-capture device (e.g., the UltraStudio Mini Recorder for Thunderbolt; Blackmagic Design Pty. Ltd.). In addition, we are currently developing a hardware implementation of the online classification and displaying system on a field-programmable gate array (FPGA), which is an integrated circuit designed to be configured by a customer or a designer after manufacturing \cite{Koide2014}. This will allow the manufacturing of a pocket-sized device that receives digital video signals, classifies video frames, and directly outputs the results to the monitor without using desktop or laptop computers.

\begin{figure}[tb]
\centering
\includegraphics[width=\linewidth]{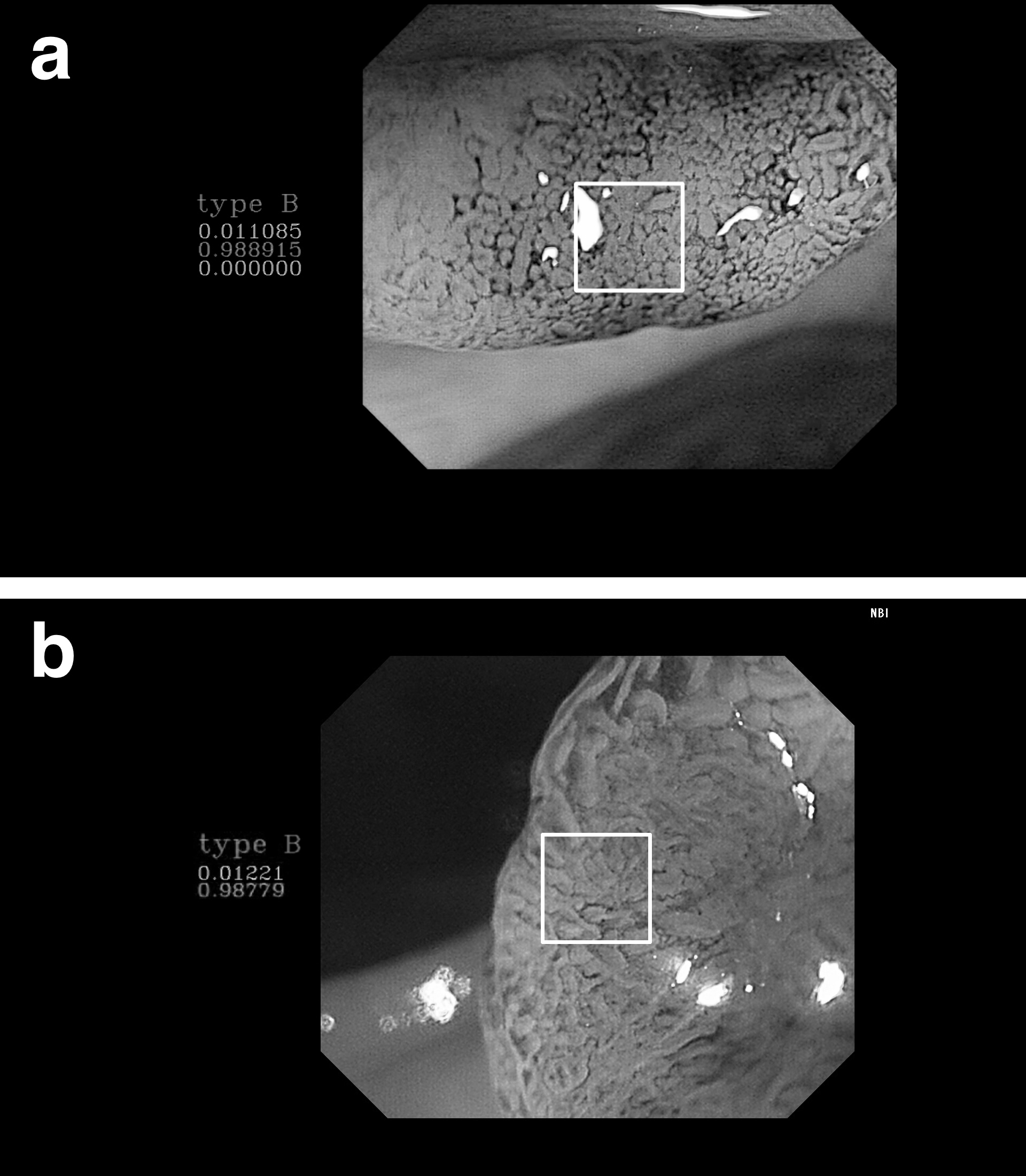}
\caption{Screen shots of the developed video frame classification system. The white square is the ROI to be classified. The results are superimposed on a video frame of (a) the endoscopic system EVIS LUCERA (shown in Figure 2) and (b) the endoscopic system EVIS LUCERA ELITE (shown in Figure 3). The frame size in both cases is 1920 $\times$ 1080 pixels. The sizes of the endoscopic video stream are (a) 1000 $\times$ 870 pixels and (b) 1156 $\times$ 1006 pixels. Note that the black background area is used to show information from the endoscopic systems such as the date, time, ID numbers, and video frame snapshots, which a have not been shown here because of confidentiality reasons.}
\label{fig:screenshot}	
\end{figure}
Figure \ref{fig:screenshot} shows sample screen shots of a monitor displaying the results of the developed CAD system. The size of the captured video frame in which the endoscopic video stream is displayed is full HD (1980 $\times$ 1080 pixels). In each video frame, an ROI patch of 200 $\times$ 200 pixels (shown as white squares in Figure \ref{fig:screenshot}) at the center of the endoscopic video stream is trimmed and classified by the pretrained SVM classifier. The classification result is shown on the left side of the endoscopic video frame. In Figure \ref{fig:screenshot}(a), the result of the three-category classification for this video frame is shown as “type B,” which is the category label given by the classifier, followed by the three probabilities of each category: 1.1\% for type A, 98.9\% for type B, and no probability for type C. Therefore, our system provides an objective measure indicating that this frame is type B with 98\% confidence, while it may be type A with probability of 1.1\%. In Figure \ref{fig:screenshot}(b), the result of the two-category classification (type A or not) is shown as “type B” (which means “not type A”) with a probability of 98.8\%, and there is still a probability of 1.2\% that this frame is type A.

The current system processes frame by frame; therefore, the probability results displayed on the monitor could become too unstable for endoscopists to determine the system output. This could be because of the incorrect classification results resulting from motion blur, out of focus images, or color bleeding. To make the classification results stable, we proposed the following two methods: a stable labeling method that could suppress frequent label changes \cite{Hirakawa2013EMBC} and a smoothing method that smoothens the probabilistic curves \cite{Hirakawa2013ICIP,Hirakawa2016}. These methods will be included in our system in future.

\begin{figure}[tb]
\centering
\includegraphics[width=\linewidth]{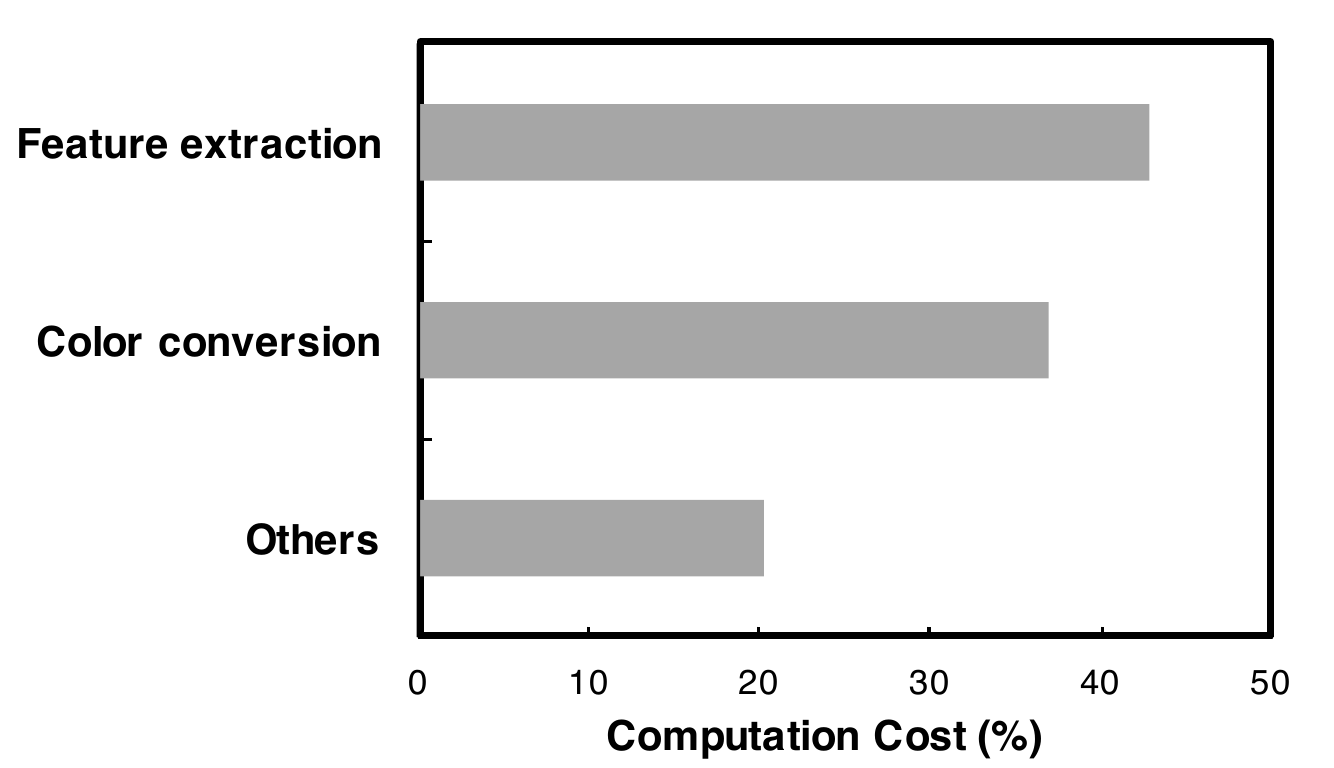}
\caption{Computational cost per frame.}
\label{fig:computationalcost}
\end{figure}
The throughput of the developed system and the endoscopic video stream are approximately about 20 fps (approximately 50 ms per frame) \cite{Kominami2016} and 30 fps, respectively. Figure \ref{fig:computationalcost} shows the relative computational cost per frame. We can see that the SIFT feature extraction and the color conversion from YUV422 to RGB require the bulk of the computation time. Further optimization of these processes needs to be done to achieve a better system throughput of up to 30fps. This is also our future work. However, the current throughput of 20 fps is high enough for normal clinical use. Note that, there is a tradeoff between the ROI rectangle size, classification performance, and processing time. If the ROI size is too small, the classification performance would decrease because of the insufficient number of SIFT features being extracted. However, classifying ROIs of larger sizes makes the classification results reliable even though the processing time increases. The current ROI size of 200 $\times$ 200 pixels is an acceptable compromise for the current result.

Apart from the abovementioned engineering aspect of the developed CAD system described above, the medical significance of our system is also important. Herein, we refer to statements concerning the real-time endoscopic assessment of the histology of diminutive (≤5 mm) colorectal polyps published by the Preservation and Incorporation of Valuable Endoscopic Innovations (PIVI) committee of the American Society for Gastrointestinal Endoscopy \cite{Rex2011}. One of the two PIVI recommendations states, “in order for a technology to be used to guide the decision to leave suspected rectosigmoid hyperplastic polyps ≤5 mm in size in place (without resection), the technology should provide ≥90\% negative predictive value ... for adenomatous histology” [26]. In other words, in the context of our system, more than 90\% of lesions diagnosed as “type A” by community endoscopists using our CAD system should histologically be non- neoplastic lesions. On the basis of this recommendation, our developed system was introduced in actual endoscopic examinations in the hospital and was evaluated for its medical significance. Details can be found in Kominami et al. \cite{Kominami2016}; however, we highlight their results here. These clinical case studies were conducted for six months (between October 2014 and March 2015). Endoscopists used our developed system to classify video frames of colon tumors into two categories, i.e., type A (non-neoplastic lesions) or types B and C3 (neoplastic lesions). Classification probabilities obtained from our developed system were evaluated in concordance with the two diagnostic results: endoscopic and histological diagnoses. For the concordance with the endoscopic diagnosis, the concordance rate was 96.6\% with a kappa static value of 0.93 and a 95\% confidence interval of 0.89--1.00. For the concordance with histology, Kominami et al. performed the Mann–Whitney U test and obtained an accuracy of 93.2\% (sensitivity: 93.0\%, specificity: 93.3\%, positive predictive value: 93.0\%, and negative predictive value: 93.3\%). Thus, these results show that nonexpert endoscopists were able to diagnose colon tumors with an accuracy sufficient to satisfy the PIVI requirement using our system.

While the endoscopists concluded that our CAD system may satisfy the PIVI recommendations, they also said that “further development of our real-time image recognition system ... and additional studies aimed at assessing whether community endoscopists may successfully meet both PIVI thresholds are needed” \cite{Kominami2016}. The motivation to publish the current paper on the system development arises from the necessity to further develop the system for clinical studies. It has been reported \cite{Ladabaum2013,Rastogi2014,Patel2016} that the performance of endoscopists is sufficient to achieve the PIVI requirement but only with a prior training module. These reports highlight the importance of the studies on the performance of non-expert endoscopists in actual clinical examinations [36], and on the necessity of training for maintaining the endoscopists’ performance \cite{Ladabaum2013,Patel2016}. We believe that the our CAD system will be useful in future clinical studies and in training and assessing the skills of endoscopists.

\section*{Conclusions}
We developed a real-time colorectal tumor classification system that provides a real-time objective measure of the status of colon tumors to endoscopists during examinations. This system was built in such a way that no modifications to the actual endoscopic systems being used in hospitals are necessary. A six-month- long clinical case study using the developed system for actual endoscopic examinations demonstrated that our system is mobile in the hospital, a processing speed of 20 fps is sufficient for examinations, and the system is medically significant from the viewpoint of the PIVI recommendations. Our future work will include making the system faster, more compact, and more user-friendly so that the system could be used by community endoscopists.

\begin{acknowledgements}
This work was supported by the JSPS Grant-in-Aid for JSPS Fellows Grant Number JP14J00223 and the Scientific Research Grant Numbers JP26280015. The authors would like to thank Misato Kawakami and Shoji Sonoyama, a former developer, for implementing the initial online recognition code with OpenCV on Windows OS and experimental SSE implementations, and Takumi Okamoto for advising us on the arrangement of the DeckLink video capture devices.
\end{acknowledgements}


\begin{thebibliography}{10}

\bibitem{Takemura2012}
Yoshito Takemura, Shigeto Yoshida, Shinji Tanaka, Rie Kawase, Keiichi Onji,
  Shiro Oka, Toru Tamaki, Bisser Raytchev, Kazufumi Kaneda, Masaharu Yoshihara,
  and Kazuaki Chayama.
\newblock Computer-aided system for predicting the histology of colorectal
  tumors by using narrow-band imaging magnifying colonoscopy (with video).
\newblock {\em Gastrointest Endosc}, 75(1):179--85, Jan 2012.

\bibitem{Gono2004}
Kazuhiro Gono, Takashi Obi, Masahiro Yamaguchi, Nagaaki Ohyama, Hirohisa
  Machida, Yasushi Sano, Shigeaki Yoshida, Yasuo Hamamoto, and Takao Endo.
\newblock Appearance of enhanced tissue features in narrow-band endoscopic
  imaging.
\newblock {\em J Biomed Opt}, 9(3):568--77, 2004.

\bibitem{Machida2004}
H~Machida, Y~Sano, Y~Hamamoto, M~Muto, T~Kozu, H~Tajiri, and S~Yoshida.
\newblock Narrow-band imaging in the diagnosis of colorectal mucosal lesions: a
  pilot study.
\newblock {\em Endoscopy}, 36(12):1094--8, Dec 2004.

\bibitem{Sano2006}
Yasushi Sano, Takahiro Horimatsu, Kuang~I. Fu, Atsushi Katagiri, Manabu Muto,
  and Hideki Ishikawa.
\newblock Magnifying observation of microvascular architecture of colorectal
  lesions using a narrow-band imaging system.
\newblock {\em Digestive Endoscopy}, 18:S44--S51, 2006.

\bibitem{Oba2010}
Sayaka Oba, Shinji Tanaka, Shiro Oka, Hiroyuki Kanao, Sigeto Yoshida, Fumio
  Shimamoto, and Kazuaki Chayama.
\newblock Characterization of colorectal tumors using narrow-band imaging
  magnification: combined diagnosis with both pit pattern and microvessel
  features.
\newblock {\em Scand J Gastroenterol}, 45(9):1084--92, Sep 2010.

\bibitem{Kanao2009}
Hiroyuki Kanao, Shinji Tanaka, Shiro Oka, Mayuko Hirata, Shigeto Yoshida, and
  Kazuaki Chayama.
\newblock Narrow-band imaging magnification predicts the histology and invasion
  depth of colorectal tumors.
\newblock {\em Gastrointestinal Endoscopy}, 69(3, Part 2):631 -- 636, 2009.

\bibitem{Meining2004}
A~Meining, T~R{\"o}sch, R~Kiesslich, M~Muders, F~Sax, and W~Heldwein.
\newblock Inter- and intra-observer variability of magnification
  chromoendoscopy for detecting specialized intestinal metaplasia at the
  gastroesophageal junction.
\newblock {\em Endoscopy}, 36(2):160---164, February 2004.

\bibitem{Mayinger2006}
Brigitte Mayinger, Yurdag{\"u}l Oezturk, Manfred Stolte, Gerhard Faller,
  Johannes Benninger, Dieter Schwab, Juergen Maiss, Eckhart~G Hahn, and Steffen
  Muehldorfer.
\newblock Evaluation of sensitivity and inter- and intra-observer variability
  in the detection of intestinal metaplasia and dysplasia in barrett's
  esophagus with enhanced magnification endoscopy.
\newblock {\em Scand J Gastroenterol}, 41(3):349--56, Mar 2006.

\bibitem{Tamaki2013MedIA}
Toru Tamaki, Junki Yoshimuta, Misato Kawakami, Bisser Raytchev, Kazufumi
  Kaneda, Shigeto Yoshida, Yoshito Takemura, Keiichi Onji, Rie Miyaki, and
  Shinji Tanaka.
\newblock Computer-aided colorectal tumor classification in {NBI} endoscopy
  using local features.
\newblock {\em Medical Image Analysis}, 17(1):78 -- 100, 2013.

\bibitem{Csurka2004}
Gabriella Csurka, Christopher Dance, Lixin Fan, Jutta Willamowski, and
  C{\'e}dric Bray.
\newblock Visual categorization with bags of keypoints.
\newblock In {\em Workshop on statistical learning in computer vision, ECCV},
  volume~1, pages 1--2. Prague, 2004.

\bibitem{Nowak2006}
Eric Nowak, Fr{\'e}d{\'e}ric Jurie, and Bill Triggs.
\newblock {\em Sampling Strategies for Bag-of-Features Image Classification},
  pages 490--503.
\newblock Springer Berlin Heidelberg, Berlin, Heidelberg, 2006.

\bibitem{Lazebnik2006}
S.~Lazebnik, C.~Schmid, and J.~Ponce.
\newblock Beyond bags of features: Spatial pyramid matching for recognizing
  natural scene categories.
\newblock In {\em 2006 IEEE Computer Society Conference on Computer Vision and
  Pattern Recognition (CVPR'06)}, volume~2, pages 2169--2178, 2006.

\bibitem{Lowe1999}
D.~G. Lowe.
\newblock Object recognition from local scale-invariant features.
\newblock In {\em Computer Vision, 1999. The Proceedings of the Seventh IEEE
  International Conference on}, volume~2, pages 1150--1157 vol.2, 1999.

\bibitem{Lowe2004}
David~G. Lowe.
\newblock Distinctive image features from scale-invariant keypoints.
\newblock {\em International Journal of Computer Vision}, 60(2):91--110, 2004.

\bibitem{FeiFei2005}
L.~Fei-Fei and P.~Perona.
\newblock A bayesian hierarchical model for learning natural scene categories.
\newblock In {\em 2005 IEEE Computer Society Conference on Computer Vision and
  Pattern Recognition (CVPR'05)}, volume~2, pages 524--531 vol. 2, June 2005.

\bibitem{Jurie2005}
F.~Jurie and B.~Triggs.
\newblock Creating efficient codebooks for visual recognition.
\newblock In {\em Tenth IEEE International Conference on Computer Vision
  (ICCV'05) Volume 1}, volume~1, pages 604--610 Vol. 1, Oct 2005.

\bibitem{Herve2009}
Nicolas Herv{\'e}, Nozha Boujemaa, and Michael~E. Houle.
\newblock Document description: what works for images should also work for
  text?, 2009.

\bibitem{Nister2006}
D.~Nister and H.~Stewenius.
\newblock Scalable recognition with a vocabulary tree.
\newblock In {\em 2006 IEEE Computer Society Conference on Computer Vision and
  Pattern Recognition (CVPR'06)}, volume~2, pages 2161--2168, 2006.

\bibitem{Stenwart2008}
I.~Steinwart and A.~Christmann.
\newblock {\em Support Vector Machines}.
\newblock NewYork, Springer, first edition, 2008.

\bibitem{Scholkopf2002}
Bernhard Sch{\"o}lkopf and Alexander~J Smola.
\newblock {\em Learning with kernels: support vector machines, regularization,
  optimization, and beyond}.
\newblock MIT press, 2002.

\bibitem{Cristianini2000}
Nello Cristianini and John Shawe-Taylor.
\newblock {\em An introduction to support vector machines and other
  kernel-based learning methods}.
\newblock Cambridge university press, 2000.

\bibitem{Vapnik1998}
Vladimir~Naumovich Vapnik and Vlamimir Vapnik.
\newblock {\em Statistical learning theory}, volume~1.
\newblock Wiley New York, 1998.

\bibitem{Hirakawa2013EMBC}
T.~Hirakawa, T.~Tamaki, B.~Raytchev, K.~Kaneda, T.~Koide, S.~Yoshida,
  Y.~Kominami, T.~Matsuo, R.~Miyaki, and S.~Tanaka.
\newblock Labeling colorectal nbi zoom-videoendoscope image sequences with mrf
  and svm.
\newblock In {\em 2013 35th Annual International Conference of the IEEE
  Engineering in Medicine and Biology Society (EMBC)}, pages 4831--4834, July
  2013.

\bibitem{Hirakawa2013ICIP}
T.~Hirakawa, T.~Tamaki, B.~Raytchev, K.~Kaneda, T.~Koide, Y.~Kominami,
  R.~Miyaki, T.~Matsuo, S.~Yoshida, and S.~Tanaka.
\newblock Smoothing posterior probabilities with a particle filter of dirichlet
  distribution for stabilizing colorectal nbi endoscopy recognition.
\newblock In {\em 2013 IEEE International Conference on Image Processing},
  pages 621--625, Sept 2013.

\bibitem{Hirakawa2016}
Tsubasa Hirakawa, Toru Tamaki, Bisser Raytchev, Kazufumi Kaneda, Tetsushi
  Koide, Shigeto Yoshida, Yoko Kominami, and Shinji Tanaka.
\newblock Defocus-aware dirichlet particle filter for stable endoscopic video
  frame recognition.
\newblock {\em Artificial Intelligence in Medicine}, 68:1 -- 16, 2016.

\bibitem{Rex2011}
Douglas~K. Rex, Charles Kahi, Michael O'Brien, T.R. Levin, Heiko Pohl, Amit
  Rastogi, Larry Burgart, Tom Imperiale, Uri Ladabaum, Jonathan Cohen, and
  David~A. Lieberman.
\newblock The american society for gastrointestinal endoscopy \{PIVI\}
  (preservation and incorporation of valuable endoscopic innovations) on
  real-time endoscopic assessment of the histology of diminutive colorectal
  polyps.
\newblock {\em Gastrointestinal Endoscopy}, 73(3):419 -- 422, 2011.

\bibitem{Bradski2000}
Gary Bradski et~al.
\newblock The opencv library.
\newblock 2000.

\bibitem{Vedaldi2010}
Andrea Vedaldi and Brian Fulkerson.
\newblock Vlfeat: An open and portable library of computer vision algorithms.
\newblock In {\em Proceedings of the 18th ACM international conference on
  Multimedia}, pages 1469--1472. ACM, 2010.

\bibitem{Chang2011}
Chih-Chung Chang and Chih-Jen Lin.
\newblock Libsvm: A library for support vector machines.
\newblock {\em ACM Trans. Intell. Syst. Technol.}, 2(3):27:1--27:27, May 2011.

\bibitem{Sonoyama2015}
S.~Sonoyama, T.~Hirakawa, T.~Tamaki, T.~Kurita, B.~Raytchev, K.~Kaneda,
  T.~Koide, S.~Yoshida, Y.~Kominami, and S.~Tanaka.
\newblock Transfer learning for bag-of-visual words approach to nbi endoscopic
  image classification.
\newblock In {\em 2015 37th Annual International Conference of the IEEE
  Engineering in Medicine and Biology Society (EMBC)}, pages 785--788, Aug
  2015.

\bibitem{Sonoyama2016}
S.~Sonoyama, T.~Tamaki, T.~Hirakawa, B.~Raytchev, K.~Kaneda, T.~Koide,
  S.~Yoshida, H.~Mieno, and S.~Tanaka.
\newblock Transfer learning for endoscopic image classification.
\newblock In {\em Proceedings of The Korea-Japan joint workshop on Frontiers of
  Computer Vision, FCV2016}, 2016.

\bibitem{Koide2014}
T.~Koide, A.~T. Hoang, T.~Okamoto, S.~Shigemi, T.~Mishima, T.~Tamaki,
  B.~Raytchev, K.~Kaneda, Y.~Kominami, R.~Miyaki, T.~Matsuo, S.~Yoshida, and
  S.~Tanaka.
\newblock Fpga implementation of type identifier for colorectal endoscopie
  images with nbi magnification.
\newblock In {\em Circuits and Systems (APCCAS), 2014 IEEE Asia Pacific
  Conference on}, pages 651--654, Nov 2014.

\bibitem{Kominami2016}
Yoko Kominami, Shigeto Yoshida, Shinji Tanaka, Yoji Sanomura, Tsubasa Hirakawa,
  Bisser Raytchev, Toru Tamaki, Tetsusi Koide, Kazufumi Kaneda, and Kazuaki
  Chayama.
\newblock Computer-aided diagnosis of colorectal polyp histology by using a
  real-time image recognition system and narrow-band imaging magnifying
  colonoscopy.
\newblock {\em Gastrointestinal Endoscopy}, 83(3):643 -- 649, 2016.

\bibitem{Ladabaum2013}
Uri Ladabaum, Ann Fioritto, Aya Mitani, Manisha Desai, Jane~P. Kim, Douglas~K.
  Rex, Thomas Imperiale, and Naresh Gunaratnam.
\newblock Real-time optical biopsy of colon polyps with narrow band imaging in
  community practice does not yet meet key thresholds for clinical decisions.
\newblock {\em Gastroenterology}, 144(1):81 -- 91, 2013.

\bibitem{Rastogi2014}
Amit Rastogi, Deepthi~S. Rao, Neil Gupta, Scott~W. Grisolano, Daniel~C.
  Buckles, Elena Sidorenko, John Bonino, Takahisa Matsuda, Evelien Dekker,
  Tonya Kaltenbach, Rajvinder Singh, Sachin Wani, Prateek Sharma, Mojtaba~S.
  Olyaee, Ajay Bansal, and James~E. East.
\newblock Impact of a computer-based teaching module on characterization of
  diminutive colon polyps by using narrow-band imaging by non-experts in
  academic and community practice: a video-based study.
\newblock {\em Gastrointestinal Endoscopy}, 79(3):390 -- 398, 2014.

\bibitem{Patel2016}
Swati~G. Patel, Philip Schoenfeld, Hyungjin~Myra Kim, Emily~K. Ward, Ajay
  Bansal, Yeonil Kim, Lindsay Hosford, Aimee Myers, Stephanie Foster, Jenna
  Craft, Samuel Shopinski, Robert~H. Wilson, Dennis~J. Ahnen, Amit Rastogi, and
  Sachin Wani.
\newblock Real-time characterization of diminutive colorectal polyp histology
  using narrow-band imaging: Implications for the resect and discard strategy.
\newblock {\em Gastroenterology}, 150(2):406 -- 418, 2016.

\end{thebibliography}

\end{document}